\documentclass{article}

\usepackage{arxiv}

\usepackage[utf8]{inputenc} 
\usepackage[T1]{fontenc}    
\usepackage{hyperref}       
\usepackage{url}            
\usepackage{booktabs}       
\usepackage{amsfonts}       
\usepackage{nicefrac}       
\usepackage{microtype}      
\usepackage{lipsum}
\usepackage{indentfirst}
\usepackage{graphicx}
\usepackage{multicol, multirow}
\usepackage{caption} 
\usepackage{subcaption}
\usepackage{float}
\usepackage{aliascnt}
\newaliascnt{eqfloat}{equation}
\newfloat{eqfloat}{h}{eqflts}
\floatname{eqfloat}{Equation}

\bgroup

\newcommand*{\ORGeqfloat}{}
\let\ORGeqfloat\eqfloat
\def\eqfloat{%
  \let\ORIGINALcaption\caption
  \def\caption{%
    \addtocounter{equation}{-1}%
    \ORIGINALcaption
  }%
  \ORGeqfloat
}

\title{Interpretation of smartphone-captured radiographs utilizing a deep learning-based approach}

\author{
 Hieu X. Le $^{1,\dagger}$, Phuong D. Nguyen $^{1,\dagger}$, Thang H. Nguyen $^1$, Khanh N.Q. Le $^{2,3,*}$, Thanh T. Nguyen $^{1,*}$ 
  \\
  $^1$ AI Research Laboratory, RE-THINKING COMPANY, Vietnam
  \\
  $^2$ Professional Master Program in Artificial Intelligence in Medicine,\\ College of Medicine, Taipei Medical University, Taipei City 106, Taiwan
  \\
  $^3$ Research Center for Artificial Intelligence in Medicine,\\Taipei Medical University, Taipei City 106, Taiwan
  \\
  $^\dagger$ Hieu and Phuong contributed equally to this work.
  \\
  $^*$ Corresponding authors: khanhlee@tmu.edu.tw, thanh.steve@rethinkingai.com
}
\date{}

\begin{document}
\maketitle

\begin{abstract} 
Recently, computer-aided diagnostic systems (CADs) that could automatically interpret medical images effectively have been the emerging subject of recent academic attention. For radiographs, several deep learning-based systems or models have been developed to study the multi-label diseases recognition tasks. However, none of them have been trained to work on smartphone-captured chest radiographs. In this study, we proposed a system that comprises a sequence of deep learning-based neural networks trained on the newly released CheXphoto dataset to tackle this issue. The proposed approach achieved promising results of 0.684 in AUC and 0.699 in average F1 score. To the best of our knowledge, this is the first published study that showed to be capable of processing smartphone-captured radiographs.  

\end{abstract}


\section{Introduction}
\parindent=2em
In the field of medical imaging, chest radiographs, or X-rays remain as the gold standard for interpreting lung conditions of one and play an important role in clinical care treatment. Recent years witnessed the rising remarkable success of Artificial intelligence (AI) technology in various fields such as computer vision or health services. In detection of diseases in medical images, especially in radiographs, AI-based systems have proven to be powerful tools that can handle medical challenges quickly and cheaply and thereby can significantly improve diagnostics quality and ultimately treat the disease.

For examples, detection of skin cancers has been enabled by a vast number of accurate deep learning studies in 2019 such as \cite{skin1} \cite{skin2} or \cite{to:hal-02335240}. Mammography, which is usually used to detect breast cancer has been the interest of such deep learning studies \cite{breast1} \cite{breast2}. A recent advanced study has also been conducted on the use of deep learning to identify Appendicitis using videos that contain CT scans\cite{ct_scans} . For radiographs, scientists also applied deep learning to detect particular conditions of lung health, such as pneumonia or consolidation, etc.… Merely, Deep Learning has proven its efficiency in a recent study to generate new synthesis data for training \cite{augmentation}. Some works even lead to the conclusion that AI-based systems can suppress the performance of normal medical doctors or qualified experts in diseases detection \cite{outperform1} \cite{outperform2}.

The clinical multi-centre has moreover provided several datasets with a variety of different diseases, serving the study of the ability to accurately diagnose including the lung. Chest x-ray image of 14 diseases was created by the National Institutes of Health Clinical Center with 112,120 frontal position radiographs of 30,805 patients \cite{wang2017chestx}. Indiana dataset provided 7470 frontal and lateral chest x-ray images of common labelling disease from the Indiana University School of Medicine \cite{demner2016preparing}. Tuberculosis image dataset from another multi-centre with normal case and disease case \cite{jaeger2014two} and lung nodule includes 100 malignant and 54 benign \cite{shiraishi2000development}. With uncertainty label, cheXpert dataset contains 224,316 images of 65,240 patients that collected from 2002 to 2017 at Stanford Hospital \cite{irvin2019chexpert}. Continuity, 10,507 photo x-ray images of 3,000 unique patients for training set that get better insight into developing detection models with high accuracy being captured by smartphones \cite{phillips2020chexphoto}.  

Due to the constant increase of large-scale dataset, several studies have significantly proposed a lot of models to classify and detect potential disease candidates, especially lung-related such as pneumonia, tuberculosis, pleural effusion, etc. Recently, mRMR algorithm was proposed for applying pneumonia classification based on different architectures (AlexNet, VGG-16, VGG-19) with high accuracy 99.41\% \cite{tougaccar2020deep}. CheXpert dataset of uncertainty labelling disease achieved 0.889 of AUROC when applied to DenseNet-121 \cite{irvin2019chexpert}. Another study also applied DenseNet-121 that showed 0.940 of AUROC of five selected diseases \cite{pham2019interpreting}, and predicted 0.90 of AUC for effusion disease \cite{allaouzi2019novel}. Two batch sizes were analyzed via different architectures, in which ResNet-50, ResNet-101, ResNet-152, and DenseNet-169 have shown highest performance when compared with CheXpert baseline for pleural effusion \cite{bressem2020comparing}.

Meanwhile, with target lung diseases we focus including pneumothorax, effusion, pneumonia, atelectasis. Guendel et al. has achieved state-of-art-result with high AUC value 0.846,0.828,0.767, 0.731, especially that was deployed on chest X-ray14 dataset \cite{guendel2018learning}. Applied view position, gender and age as a useful metadata that investigated better insight into other factors outside chest X-ray \cite{baltruschat2019comparison}. However, the evaluation of posteroanterior and anteroposterior is necessary for model training to tackle different view positions \cite{rubin2018large}, especially anteroposterior might recognize pleural effusion disease.  Recently, a group of research scientists also proposed a new way to remove foreign objects, such as pace-makers or electric wires from radiographs, using a subset of the CheXphoto dataset \cite{remove-object}.

The wide access to smartphones, together with lack of qualified radiologists in remote areas have raised the need for another study for using smartphone-captured radiographs to provide automatic, accurate and distant diagnosis of lung diseases for patients in remote areas. Careful examination of published studies reveals a fact that although a large amount of studies has been carried out for lung disease recognition, none of them have been conducted on radiographs that were captured by smartphone cameras. This observation motivates us to build a method to enable lung disease detection via radiographs that are captured by smartphone cameras. The results obtained in this study are promising and may serve as a facilitator and baseline for other research in the future. In addition, for the ease of wide access, we also plan to deliver the product of this study as a mobile application.

This study yielded several key contributions, stated as following: Firstly, we proposed the first and novel system that can work with radiographs captured by smartphone cameras. Compared to other existing studies, we do not require the isolation of radiographs and removal of background objects. Secondly, we proposed a proper way to tune the classification thresholds for improving the model performance. Our system also obtained satisfactory results with...

The remainder of the paper is organized as follows: In section 2, we described the used dataset and the general pipeline. Subcomponents of the systems are also explained with detailed training strategies and parameters. Achievements are presented in section 3, where we also provided evaluative metrics of the systems. In section 4, we discuss key findings as long as the advantages and limitations of current methods. Finally, we concluded the study and re-emphasized our contributions in section 5. 

\section{Experiment Design}

\subsection{Dataset preparation}
The dataset used in this study is supplied by the CheXPhoto 2020 competition \cite{chexphoto}. The competition provides a training set including natural and synthetic transformations of 10,507 xrays taken from 3000 unique patients sampled from the original CheXpert dataset \cite{chexphoto}. The provided validation set which consists of 244 x-rays is taken for self-evaluation and hyperparameter tuning purposes. Visualized examples from the CheXphoto dataset is shown in figure 1.

Each labelled x-ray in the dataset includes the annotations of 14 labels : 0 for non existence, 1 for having the observation and -1 for uncertainty. Since the CheXphoto competition requires submission of 5 classes Atelectasis, Cardiomegaly, Consolidation, Edema, Pleural Effusion only, we mainly focus our approach on these considered labels. The distribution of these labels in the original training set is reviewed in table 1, while the same figures for the validation set is shown in table 2. 

\begin{table}[]
    \centering
    \begin{tabular}{|l|r|r|r|}
    \hline
    \multicolumn{1}{|c|}{Label} & \multicolumn{1}{c|}{Positive (\%)} & \multicolumn{1}{c|}{Negative (\%)} & \multicolumn{1}{c|}{Uncertain (\%)} \\ \hline
    Atelectasis                 & 1577 (15.01)                       & 7335 (69.81)                       & 1595 (15.18)                        \\ \hline
    Cardiomegaly                & 1313 (12.5)                        & 8824 (83.98)                       & 370 (3.52)                          \\ \hline
    Consolidation               & 671 (6.39)                         & 8521 (81.1)                        & 1315 (13.52)                        \\ \hline
    Edema                       & 2553 (24.3)                        & 7320 (69.67)                       & 634 (6.03)                          \\ \hline
    Pleural Effusion            & 4115 (39.16)                       & 607 (5.81)                         & 607 (5.78)                          \\ \hline
    \end{tabular}
    \caption{Distribution of submission labels in the training set}
    \label{tab: dist-training-set}
\end{table}

\begin{table}[]
\centering
\begin{tabular}{|l|l|l|}
\hline
Label            & Positive (\%) & Negative (\%) \\ \hline
Atelectasis      & 194 (79.51)   & 50 (20.49)    \\ \hline
Cardiomegaly     & 140 (57.38)   & 104 (42.62)   \\ \hline
Consolidation    & 165 (67.62)   & 79 (32.37)    \\ \hline
Edema            & 163 (66.80)   & 81 (33.20)    \\ \hline
Pleural Effusion & 190 (77.87)   & 54 (22.13)    \\ \hline

\end{tabular}
\caption{Distribution of labels in the validation set}
\end{table}

In training, uncertain labels were ignored and all null labels were assigned to negative. The training set is further divided into two smaller sets of ratio 80:20 for learning and validation purposes. Training a model for separation of radiographs from the initial images can significantly reduce the distraction sources and improve the performance of the model. To allow separation, we further annotated the rectangular bounding region surrounding the radiographs inside images. Annotations were cross validated by at least 2 annotators after being processed by one.
\begin{figure}[htp]
    \centering
    \includegraphics[width=14cm]{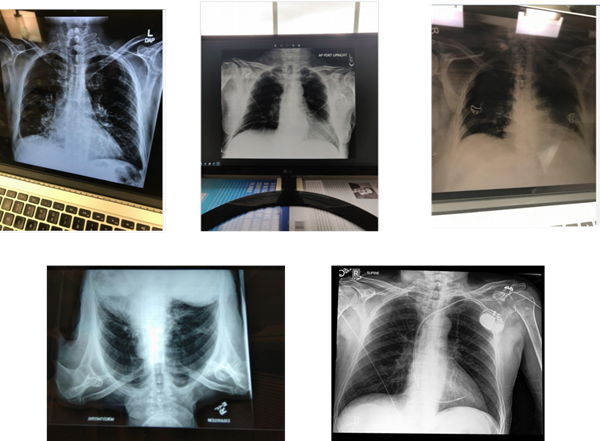}
    \caption{Examples from the CheXPhoto dataset}
    \label{fig:galaxy}
\end{figure}

\subsection{Problem formulation and study pipeline}
The objective of this research is to propose a deep learning-based approach for detection of existing lung diseases in smartphone-captured radiographs. Given an input image denoted as a matrix $X_i$, and a set of labels y = {$y_1$,$y_2$,$y_3$,...$y_t$} the proposed system is set to learn the probability of having an observation k $\subset$ y , that is:

\begin{eqfloat}
    \begin{equation}
        P(x_i , \Theta) = \frac{1}{1 + e^{-z_i}}
    \end{equation}
    \caption{Probabilities assigned for each label}
    \label{eq:MDOC}
\end{eqfloat}

Where k $\subset$ y, $x_i$ denoted the input and $z_i$ denoted the output if propagated by a network with all parameters denoted as $\Theta$.

The suggested solution includes 3 main neural network components: the radiograph localization module, the image-processing module, and the main disease detection model.

In the localization module, we made use of the widely Yolo-v3 architecture [8]. Each input image is predicted to gain a rectangle box bounding the regions containing only the radiographs. Subsequently, in the pre-processing module, the image is then normalized to a predefined length of values and size to allow consistency of the data flow. One may witness the existence of foreign objects in some particular radiographs in the dataset, thus redundant objects removal was applied as following the procedure described in \cite{remove-object}.  Finally, we proposed a deep learning model to detect certain lung diseases. Figure 1 visualized the overall pipeline of this study.

\begin{figure}[htp]
    \centering
    \includegraphics[width=12cm]{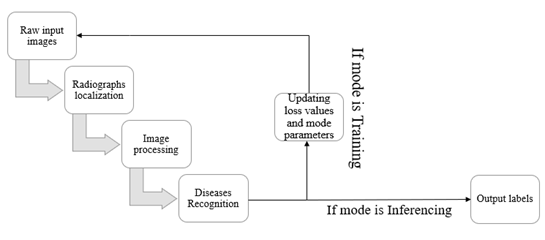}
    \caption{Pipeline of the study}
    \label{fig:galaxy}
\end{figure}

\subsection{Radiograph detection model training strategies}

\textbf{Model architecture}

In this dataset, radiographs are usually captured with different devices under various settings and thus, there may exist background or non-related objects inside each input image. It is thereby reasonable to consider only the regions contains radiographs when learning or inference, as these background objects and artifacts does not provide any information but can be observed as sources of noise and distraction for the models. In this study, we adopted the YOLOv3 architecture \cite{yolov3} to detect the bounding regions of radiographs in raw inputs.   

\textbf{Model optimization }

Optimization of the model is carried out with a batch size of 8, while the loss function used is GIoU loss. Weights were updated every iteration with the SGD Optimizer. For every iteration, we saved the best checkpoint measured by loss values. After the training process, only the checkpoint which has lowest loss values was then considered for extracting the radiographs from the input images of the test set and for evaluation on the validation dataset. We did not implement test time augmentation or any other post-processing techniques in this stage.

\textbf{Model inference }

When inferencing on the validation or the test set, we set up the batch size to 1, and disable the augmentation option. Other thresholds were set as default. After cropping out the radiographs, extracted regions were normalized to the range of [0-1] in pixel intensities and resized to a predefined size of 476x476. 

\subsection{Lung diseases recognition model training strategies}
\textbf{Model architecture }

The succeed of transfer learning techniques as long as the existence of state-of-the-art model architectures which have continuously showed an advancement in traditional image classification challenges, such as ImageNet [9] has convinced us to adopt the widely-known EfficientNet B4 architecture for this task. With a slight modification in the last fully connected layer, we have thereby changed the classification layer of the EfficientNet to fit our purposes for detecting five different classes. A visualization of the model’s head architecture is illustrated in Table 2. Transfer learning has proven its efficiency in cutting down the training time \cite{transfer1} \cite{transfer2}. Merely, a pretrained model enables the ability to leverage its convolutional base, which is trained on a large and more general dataset to provide a closer to optimal and better starting point for the current problem. We therefore adopt its pretrained weights as the initializing parameters. 

\begin{table}[]
    \centering
\begin{tabular}{|r|l|c|r|}
\hline
\multicolumn{1}{|c|}{\textbf{Input Shape}} & \multicolumn{1}{c|}{\textbf{Layer Type}} & \textbf{Bias} & \multicolumn{1}{c|}{\textbf{Output Shape}} \\ \hline
\textbf{1x1408}                            & Fully Connected Layer                    & True          & 1x512                                      \\ \hline
\textbf{1x512}                             & Batch Normalization                      & -             & 1x512                                      \\ \hline
\textbf{1x512}                             & Fully Connected Layer                    & True          & 1x5                                        \\ \hline
\textbf{1x5}                               & Sigmoid                                  & -             & 1x5                                        \\ \hline
\end{tabular}
\caption{Architecture of the model head}
\label{tab: architecture}
\end{table}

\textbf{Data Augmentation }

Conventionally, more data usually implies better performance and better generalization for the learning ability of the models. In our cases, additional or external datasets are not available. We therefore employed different data augmentation techniques to first, improve the generalization of the model, and to second, prevents overfitting. We avoid directly affecting the lung regions inside the radiographs, so that only Resizing, Horizontal Flip, Vertical Flip, Rotation and Center Crop are employed. The probabilities of using each augmentation technique are equally set to 0.5.

\textbf{Training strategies}

Defining the loss function, the optimizer and the learning scheduler is important in training. Since for each radiograph, multi labels may existed, we set the activation of the last fully connected layer to sigmoid. After sigmoid activation, the prediction score for each neuron is in the range of [0-1], we subsequently defined the classification threshold for each label to recognize its existence in the input images. The utilized loss function is Binary Cross Entropy, which is a simpler version of the conventional Cross Entropy. During training, the model’s parameters were updated utilizing the Adam optimizer with the regulation of Cosine Annealing Learning Rate scheduler. Batch size is determined in the way such that we could use the maximum capacity of the GPU’s memory. During each epoch, the weights of the model is locally saved. For inferencing, we consider the model version that had best performance when measured by the Auc-roc score. All of the models were trained on a NVIDIA RTX 1080 Ti GPU, while training codes are written in Pytorch.

\textbf{Threshold defining }

For evaluation metrics such as AUC-ROC curve, a predefined threshold is not required as it provides a general view on the overall performance of the model on a predefined set. However, metrics such as f1 score, precision and recall score do rely on the classification thresholds as they are measured using the True positive samples and the True negative samples. On the other hand, studies conducted in multi-label problems usually used 0.5 as a default threshold or manually picked up the suitable thresholds based on different experiments \cite{threshold1} \cite{threshold2}. We defined an automatic and different strategy for defining the classification thresholds. First, we assume that each label of the predictions may has different sensitivity, therefore it is reasonable to set up different classification thresholds for different labels. Observing that the ROC curves of labels may hold the key to the problem, we made use of the Youden’s index \cite{youden} to automatically pick-up a classification threshold for each label such that for that label, the subtraction of true positive rate and the false positive rate is maximized.   

\textbf{Inferencing}

For saving computing resources, all inferences were conducted on CPU only with a batch size of 1. During the inference stage, no augmentation methods should be applied.

\subsection{Evaluation metrics}
The CheXphoto competition’s official validation set were used as the official test set to assess the performance of our proposed method on the disease recognition task. For the evaluation of how well the system can perform on an independent test set, the Area under the receiver operating characteristic curve (AUC-ROC) of each label is reported together with their average figure. Precision - recall and F1-score for each label on the test set were also presented for deeper result analysis. In addition, for the radiographs detection task, we report the loss plot and the mAP.





\section{Result and discussion}
In this section, we reported attained metrics discussed in section 2.3 as well as adopted hyperparameters when using aforementioned training strategies. Figure 3 visualizes the training loss of the object detection model during training stage, including the heat map generated by Grad-CAM method to indicate the concentrated region of the model and the corresponding output probabilities. It is essential to tune the hyperparameters during training a deep learning model. To allow easier replication of the experiments inside this research, we provided the set of hyperparameter values inside table 3.

\begin{table}[h]
\centering
\begin{tabular}{|c|c|c|}
\hline
\textbf{Hyperparameter}   & \textbf{Initial value} & \textbf{Decay}             \\ \hline
\textbf{Learning rate}    & 0.003                  & By a halve every 10 epochs \\ \hline
\textbf{Batch size}       & 128                    & False                      \\ \hline
\textbf{Number of epochs} & 100                    & -                          \\ \hline
\end{tabular}
\caption{List of hyperparameters}
\label{tab: hyperparameters}
\end{table}

\begin{figure}[htp]
    \centering
    \includegraphics[width=12cm]{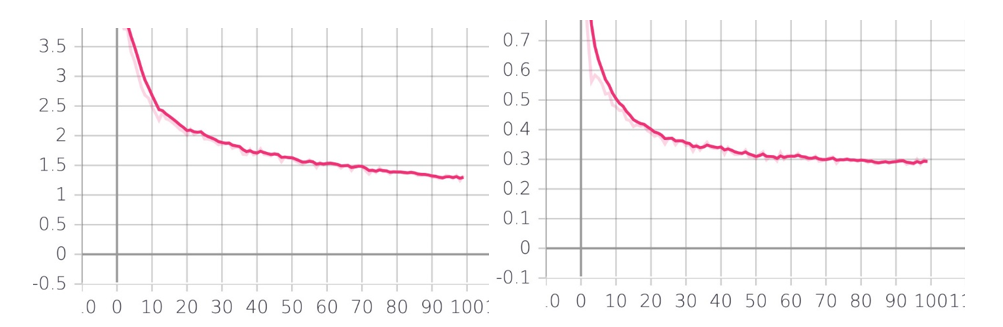}
    \caption{The GIoU loss and the object recognition loss of the YoLoV3 model}
    \label{fig:galaxy}
\end{figure}

Observed ROC curves for each label are visualized in figure 4. Other adopted metrics on the test set are presented in table 5.  
\begin{figure}[htp]
    \centering
    \includegraphics[width=10cm]{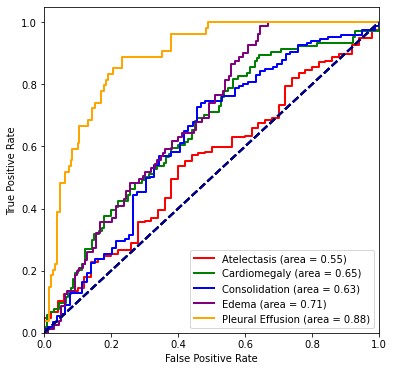}
    \caption{Obtained ROC curves for test set}
    \label{fig:galaxy}
\end{figure}

\begin{table}[h]
\centering
\begin{tabular}{|l|l|l|l|}
\hline
Label            & Precision & Recall & F1 score \\ \hline
Atelectasis      & 1.000     & 0.795  & 0.886    \\ \hline
Cardiomegaly     & 0.999     & 0.502  & 0.668    \\ \hline
Consolidation    & 1.000     & 0.676  & 0.807    \\ \hline
Edema            & 1.000     & 0.352  & 0.521    \\ \hline
Pleural Effusion & 1.000     & 0.441  & 0.612    \\ \hline
Mean value       & 1.000     & 0.553  & 0.6988   \\ \hline
\end{tabular}
\caption{ Achieved metrics for the test set}
\label{tab: metrics}
\end{table}

From Figure 3, we could observe that our proposed system is capable of solving multi-label recognition tasks at a satisfactory level. For more details, the mean Area under the curve (mean AUC) attained reaches up to 0.684. For all labels, the AUC is all more than 0.5, reflecting that the system has certain discrimination and no random guess occurred. Table 4 reveals detailed insights of the system’s predictions on the test set. We observed that precision is higher than recall for all cases.This implies that the system has high confidence when confirming with detected positive labels but is more redundant to identifying the positive labels of a sample. Label confirmed with highest accuracy is Pleural Effusion while the least accurate detected label is Edema in recall and F1 score (0.352, 0.521, respectively) because of imbalanced data. Several different augmentations can't also handle or mitigate this problem. In summary, mean values have shown recall output (0.5533) and 0.6988 for F1 score. We obtained an acceptable average accuracy of 70\% for the test set. Because of the complicated dataset, this output will constantly improve.

\textbf{Advantages and limitations}

With these key findings, we continue to assert that Deep Learning algorithms is promising for the lung disease risk recognition using the smartphone-captured radiographs. Moreover, with the highly earned precision/recall, this method is capable of ruling out/in the diseases and thus, suitable for assisting medical doctors in screening, diagnosing lung diseases. Comparison between our proposed method and other single model-based classifier trained and tested on the same data re-emphasized the importance of processing images and feeding valuable only inputs to deep learning-based classifiers. In details, we acknowledged that, whether the ability of the classifier to recognize predefined classes is good or not is really dependent on types of data fed, for instances, in this case, feeding in the radiographs separated from the background provided better results in comparison to utilizing raw input images contain both radiographs and background objects. In the future, our proposed method can be referenced or followed by researchers who seek to find deep learning-based solutions that work on smartphone-captured vision data.  

Meanwhile, we acknowledge that our proposed method has certain limitations. Noting that our solution requires training and tuning more than one model, we caution that the proposed solution should have taken more time and effort than training a single model from scratch. We also caution that this deep learning-based solution consider the vision inputs only, while in reality, radiologists and medical doctors usually take into their consideration additional metadata, such as age, gender, smoking habit, etc... To overcome this, we suggest collecting another dataset in which additional data mentioned are recorded carefully for every single case. Learning on both additional metadata-supplied dataset and images only dataset would provide a more general effective and adaptive solution which can work in both case when metadata are additionally available and when only radiographs are available.   

\textbf{Future work  }

Any single model may suffer from bias and inconsistency in remaining good AUC scores for all classes. One method to tackle this issue is to propose the use of ensemble learning. In detail, following the same training strategies, we trained the classification models with different backbones. The primary idea of ensemble learning is to use “the power of the crowd”, that is to use decisions made by several weak classifiers to obtain a less biased decision by taking the mean of probabilities or by following the major voting rules. Results are often thereby improved in comparison to using a single model. In the future, we wish to take advantage of state-of-the-art backbones such as DenseNet121, or Resnet50, to obtain a variety of classifiers. By stacking these models, we hope to improve the performance of the system.  

Telemedicine and remote diagnosis have received much attention from societies recently. For remote and rural areas, experienced and qualified radiologists are in short supplies. Therefore, in the future, we wish to develop a smartphone application that comes with a deep learning model to allow remote access for radiographs interpretations. The application can also serve as a reference for medical doctors, or inexperienced radiologists. 

\section{Conclusion}
Presented in this study was a pipeline for building an end-to-end deep learning-based system that is capable of capturing the risk of having certain observations in smartphone-captured radiographs. We proposed a novel way to deal with smartphone-captured data by using object detection models before passing to the customized classifier. We also proposed the use of Youden’s index in ROC curves for finding suitable thresholds for different classes. Promising results encouraged the aspect of using smartphone-based data, especially computer vision data in the medical imaging field.  

For better replication of all experiments, we also provided a set of training parameters as long as the loss plots of each model in the system. Loss values indicated that trained models are stable and converged. Although not achieving perfect results, the system still showed an acceptable discrimination ability for different diseases.  

On the clinical side, we hope that this work can serve as a reference for confirming one lung condition thanks to its high precision. We hope that further studies could be conducted to improve the existing method. Radiologists or clinical units may take this study into consideration for referencing ideas and saving their time and effort. With the wide availability of smartphones, we expect that these technologies can be transferred into a mobile application on all various platforms including Android or IOS, to provide easy access and usage for patients in remote areas.

\bibliographystyle{unsrt}
\bibliography{template}  


\end{document}